\documentclass[runningheads,a4paper]{llncs}

\makeatletter
\newcommand{\@chapapp}{\relax}%
\makeatother

\usepackage{amssymb}
\usepackage{graphicx}

\usepackage{color}
\usepackage[table]{xcolor} 
\usepackage{tabularx} 
\usepackage{gensymb} 
\usepackage{amssymb} 
\usepackage{bm} 
\usepackage{arydshln}
\usepackage{amsmath}
\usepackage{breqn}
\usepackage{soul}
\usepackage{epstopdf}
\usepackage{multirow}
\usepackage{graphicx}
\usepackage{caption}
\usepackage{subfig}
\usepackage{hhline}
\usepackage[normalem]{ulem}
\usepackage[super]{nth}
\usepackage[]{appendix}
\usepackage{multibib}
\newcites{appendix}{Appendix References}

\makeatletter
\newcommand\footnoteref[1]{\protected@xdef\@thefnmark{\ref{#1}}\@footnotemark}
\makeatother

\DeclareCaptionLabelFormat{andfigure}{#1~#2~\& \tablename~\thetable}

\newcommand\norm[1]{\left\lVert#1\right\rVert}

\newcolumntype{L}[1]{>{\hsize=#1\hsize\raggedright\arraybackslash}X}%
\newcolumntype{R}[1]{>{\hsize=#1\hsize\raggedleft\arraybackslash}X}%
\newcolumntype{C}[1]{>{\hsize=#1\hsize\centering\arraybackslash}X}%

\newcolumntype{R}[2]{%
    >{\adjustbox{angle=#1,lap=\width-(#2)}\bgroup}%
    l%
    <{\egroup}%
}

\begin{document}

\mainmatter  

\title{Instance Segmentation and Tracking with Cosine Embeddings and Recurrent Hourglass Networks}

\titlerunning{Cosine Embeddings with Recurrent Hourglass Networks}

%
%

\author{Christian Payer\inst{1,\thanks{This work was supported by the Austrian Science Fund (FWF): P28078-N33.}} \and
Darko \v{S}tern\inst{2} \and
Thomas Neff\inst{1} \and\\
Horst Bischof\inst{1} \and
Martin Urschler\inst{2,3}}


\authorrunning{Payer et al.}


\institute{
$^{1}$Institute of Computer Graphics and Vision, Graz University of Technology, Austria\\
$^{2}$Ludwig Boltzmann Institute for Clinical Forensic Imaging, Graz, Austria\\
$^{3}$BioTechMed-Graz, Graz, Austria
}

%
%

\maketitle

\begin{abstract}
Different to semantic segmentation, instance segmentation assigns unique labels to each individual instance of the same class.
In this work, we propose a novel recurrent fully convolutional network architecture for tracking such instance segmentations over time. 
The network architecture incorporates convolutional gated recurrent units (\mbox{ConvGRU}) into a stacked hourglass network to utilize temporal video information.
Furthermore, we train the network with a novel embedding loss based on cosine similarities, such that the network predicts unique embeddings for every instance throughout videos.
Afterwards, these embeddings are clustered among subsequent video frames to create the final tracked instance segmentations.
We evaluate the recurrent hourglass network by segmenting left ventricles in MR videos of the heart, where it outperforms a network that does not incorporate video information.
Furthermore, we show applicability of the cosine embedding loss for segmenting leaf instances on still images of plants.
Finally, we evaluate the framework for instance segmentation and tracking on six datasets of the ISBI celltracking challenge, where it shows state-of-the-art performance.
\end{abstract}

\section{Introduction}
\label{sec:intro}

\nocite{*}


Instance segmentation plays an important role in biomedical imaging tasks like cell migration, but also in computer vision based tasks like scene understanding.
It is considerably more difficult than semantic segmentation (e.g.,~\cite{Payer2018}), since instance segmentation does not only assign class labels to pixels, but also distinguishes between instances within each class, e.g., each individual person on an image from a surveillance camera is assigned a unique ID.


Mainly due to the high performance of the U-Net~\cite{Ronneberger2015}, semantic segmentation has been successfully used as a first step in medical instance segmentation tasks, e.g., cell tracking.
However, for instances to be separated as connected components during postprocessing, borders of instances have to be treated with special care.
In the computer vision community, many methods for instance segmentation have in common that they solely segment one instance at a time.
In~\cite{He2017}, all instances are first detected and independently segmented, while in~\cite{Ren2017}, recurrent networks are used to memorize which instances were already segmented.
Segmenting solely one instance at a time can be problematic when hundreds of instances are visible in the image, as often is the case with e.g., cell instance segmentation.
Recent methods are segmenting each instance simultaneously, by predicting embeddings for all pixels at once~\cite{Newell2017,Kong2017}.
These embeddings have similar values within an instance, but differ among instances.
In the task of cell segmentation and tracking, temporal information is an important cue to establish coherence between frames, thus preserving instances throughout videos.
Despite improvements of instance segmentation using embeddings, to the best of our knowledge, combining them with temporal information for tracking instance segmentations has not been presented.

\begin{figure}[t]
\centering
\includegraphics[width=\textwidth]{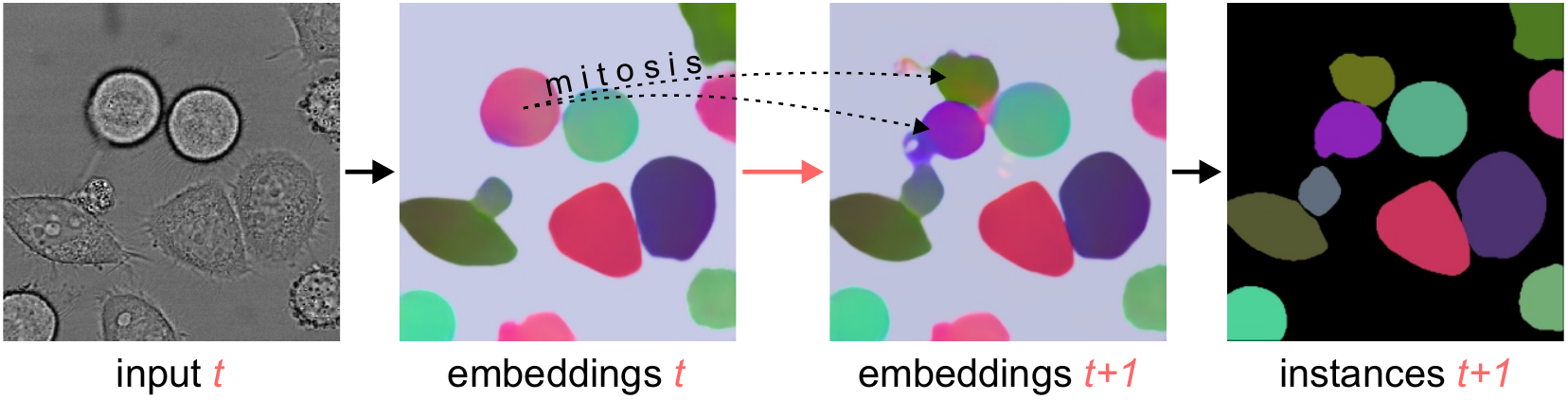}
\caption{Overview of our proposed framework showing input image, propagation of cosine embeddings from frame $t$ to frame $t+1$ (three randomly chosen embedding dimensions as RGB channels), and resulting clustered instances.}
\label{fig:overview}
\end{figure}

In this paper, we propose to use recurrent fully convolutional networks for embedding-based instance segmentation and tracking.
To memorize temporal information, we integrate convolutional gated recurrent units (\mbox{ConvGRU}~\cite{Ballas2015}) into a stacked hourglass network~\cite{Newell2016}. 
Furthermore, we use a novel embedding loss based on cosine similarities, where we exploit the four color map theorem~\cite{Appel1976}, by requiring only neighboring instances to have different embeddings.

\section{Instance Segmentation and Tracking}
\label{sec:instance_tracking}

Figure~\ref{fig:overview} shows our proposed framework on a cell instance segmentation and tracking example.
To distinguish cell instances, they are represented as embeddings at different time points.
By representing temporal sequences of embeddings in a recurrent hourglass network, a predictor can be learnt from the data, which allows tracking of embeddings also in the case of mitosis events.
To finally generate instance segmentations, clustering of the predicted embeddings is performed.



\subsection{Recurrent Stacked Hourglass Network}
\label{subsec:network}

\begin{figure}[t]
\centering
\includegraphics[width=\textwidth]{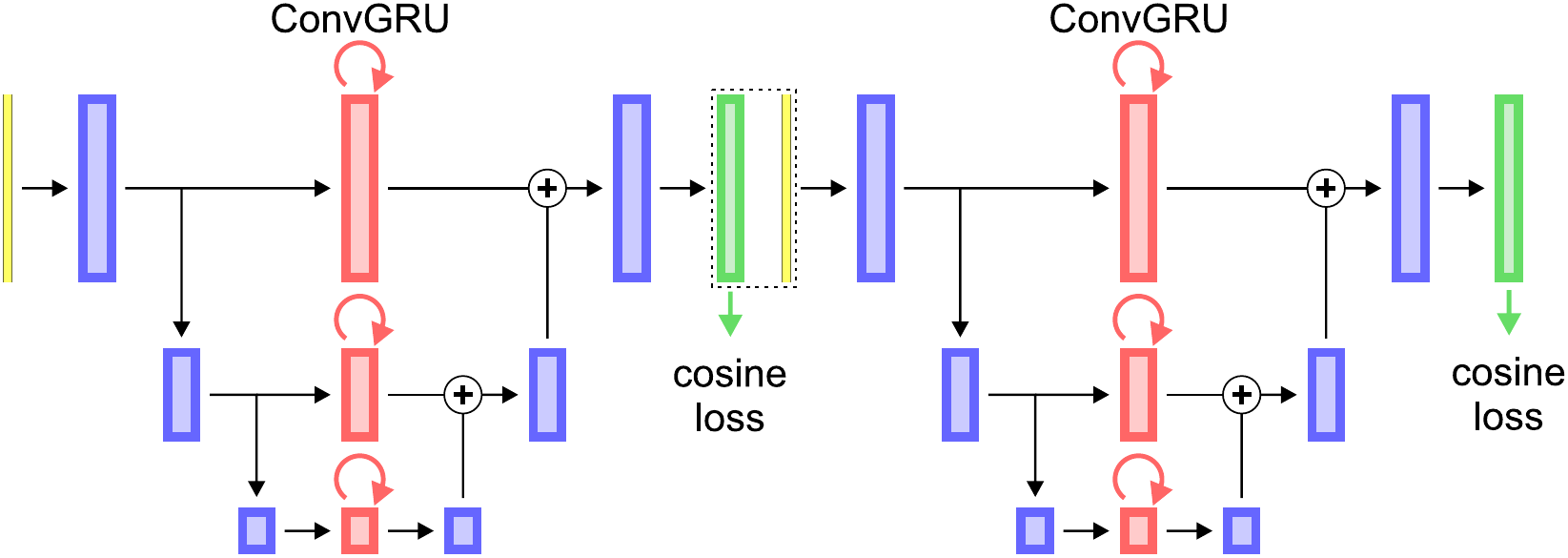}
\caption{Overview of the recurrent stacked hourglass network with two hourglasses and three levels.
Yellow bars: input; blue boxes: convolutions; red boxes: \mbox{ConvGRU}; dashed black box: concatenation; green boxes: embeddings.}
\label{fig:architecture}
\end{figure}

We modify the stacked hourglass architecture~\cite{Newell2016} by integrating \mbox{ConvGRU}~\cite{Ballas2015} to propagate temporal information, as shown in Fig.~\ref{fig:architecture}.
Differently from the original stacked hourglass network, we use single convolution layers with $3\times3$ filters and 64 outputs for all blocks in the contracting and expanding paths, while we use \mbox{ConvGRU} with $3\times3$ filters and 64 outputs in between paths.
As proposed by~\cite{Newell2016}, we also stack two hourglasses in a row to improve network predictions.
Therefore, we concatenate the output of the first hourglass with the input image to use it as input for the second hourglass.
We apply the loss function on the outputs of both hourglasses, while we only use the outputs of the second hourglass for the clustering of embeddings.

\subsection{Cosine Embedding Loss}
\label{subsec:loss}

\newcommand{\setinst}{\mathbb{I}}
\newcommand{\setseg}{\mathbb{S}}
\newcommand{\setsegi}{\setseg^{(i)}}
\newcommand{\setsegj}{\setseg^{(j)}}
\newcommand{\setsegbac}{\setseg^{\textit{bac}}}
\newcommand{\neighseg}{\mathbb{N}}
\newcommand{\neighsegi}{\neighseg^{(i)}}
\newcommand{\neighsegbac}{\neighseg^{\textit{bac}}}
\newcommand{\emb}{\vec{e}}
\newcommand{\meanemb}{\vec{\bar{e}}}
\newcommand{\meanembi}{\meanemb^{(i)}}
\newcommand{\loss}{L}

We let the network predict a $d$-dimensional embedding vector $\emb_p\in\mathbb{R}^d$ for each pixel~$p$ of the image. 
To separate instances $i\in\setinst$, firstly, embeddings of pixels $p\in\setsegi$ belonging to the same instance $i$ need to be similar, and secondly, embeddings of~$\setsegi$ need to be dissimilar to embeddings of pixels $p\in\setsegj$ of other instances $j\neq i$. 
Here, we treat background as an independent instance.
Following from the four color map theorem~\cite{Appel1976}, only neighboring instances need to have different embeddings.
Thus, we relax the need of dissimilarity between different instances only to the neighboring ones, i.e., $\neighsegi = \bigcup_j\setsegj$ for all instances $j\neq i$ within pixel-wise distance $r_{\mathbb{N}}$ to instance $i$.
This relaxation simplifies the problem by assigning only a limited number of different embeddings to a possibly large number of different instances.

We compare two embeddings with the cosine similarity
\begin{equation}
\text{cos}(\vec{e}_1,\vec{e}_2)=\frac{\vec{e}_1\cdot\vec{e}_2}{\norm{\vec{e}_1}\norm{\vec{e}_2}},
\end{equation}
which ranges from~$-1$ to~1, while~$-1$ indicates the vectors have the opposite, $0$ orthogonal, and $1$ the same direction.
We define the cosine embedding loss as
\begin{equation}
\loss=\frac{1}{|\setinst|}\sum_{i\in\setinst}\left(1-\frac{1}{|\setsegi|}\sum_{p\in\setsegi}{\text{cos}(\meanembi,\emb_p)}\right) + \left(\frac{1}{|\neighsegi|}\sum_{p\in\neighsegi}{\text{cos}(\meanembi,\emb_p)^2}\right),
\end{equation}
where the mean embedding of instance $i$ is defined as
$\meanembi=\frac{1}{|\setsegi|}\sum_{p\in \setsegi}{\emb_p}$.
By minimizing~$\loss$, the first term urges embeddings~$\emb_p$ of pixels $p\in\setsegi$ to have the same direction as the mean~$\meanembi$, which is the case when $\text{cos}(\meanembi,\emb_p)\approx1$, while the second term pushes embeddings~$\emb_p$ of pixels $p\in\neighsegi$ to be orthogonal to the mean~$\meanembi$, i.e., $\text{cos}(\meanembi,\emb_p)\approx0$.


\subsection{Clustering of Embeddings}
\label{subsec:clustering}

To get the final segmentations from the predicted embeddings, individual groups of embeddings that describe different instances need to be identified.
As the number of instances is not known, we perform this grouping with the clustering algorithm HDBSCAN~\cite{Campello2015} that estimates the number of clusters automatically.
For each dataset, two HDBSCAN parameters have to be adjusted: minimal points $m_{\text{pts}}$ and minimal cluster size $m_{\text{clSize}}$.
To simplify clustering and still be able to detect splitting of instances, we cluster only overlapping pairs of consecutive frames at a time. 
Since our embedding loss allows same embeddings for different instances that are far apart, we use both image coordinates and value of the embeddings as data points for the clustering algorithm.
After identifying the embedding clusters with HDBSCAN and filtering clusters that are smaller than $t_{\text{size}}$, the final segmented instances for each frame pair are obtained.

For merging the segmented instances in overlapping frame pairs, we identify same instances by the highest intersection over union (IoU) between each segmented instance in the overlapping frame.
The resulting segmentations are then upsampled back to the original image size, generating the final segmented and tracked instances.

\section{Experimental Setup and Results}
\label{sec:setup}

We train the networks with TensorFlow\footnote{\url{https://www.tensorflow.org/}} and perform on-the-fly data augmentation with SimpleITK\footnote{\url{http://www.simpleitk.org/}}.
We use hourglass networks with seven levels and an input size of $256\times256$, while we scale the input images to fit.
All recurrent networks are trained on sequences of ten frames.
We refer to the supplementary material for individual training and augmentation parameters, as well as individual values of parameter described in Section~\ref{sec:instance_tracking}.

\noindent\textbf{Left Ventricle Segmentation:}
To show that our proposed recurrent stacked hourglass network is able to incorporate temporal information, we perform semantic segmentation on videos of short-axis MR slices of the heart from the left ventricle segmentation challenge~\cite{Suinesiaputra2014}.
We compare the recurrent network with a non-recurrent version, where we replace each \mbox{ConvGRU} with a convolution layer to keep the network complexity the same.
Since outer slices do not contain parts of the left ventricle, the networks are evaluated on the three central slices that contain both left ventricle myocardium and blood cavity (see Fig.~\ref{fig:heart}). 
We train the networks with a softmax cross entropy loss to segment three labels, i.e., background, myocardium, and blood cavity.
We use a three-fold cross-validation setup, where we randomly split datasets of 96 patients into three equally sized folds.
Table~\ref{tbl:heart} shows the IoU for our internal cross-validation of both recurrent and non-recurrent stacked hourglass networks.

\noindent\textbf{Leaf Instance Segmentation:}
We show that the cosine embedding loss and the subsequent clustering are suitable for instance segmentation without temporal information, by evaluating on the A1 dataset of the CVPPP challenge for segmenting individual plant leaves~\cite{Minervini2016} (see Fig.~\ref{fig:plant}).
We use the non-recurrent version of the proposed network from the previous experiment to predict embeddings with 32 dimensions.
Consequently, the clustering is also performed on single images.
As we were not able to provide results on the challenge test set in time before finalizing this paper, we report results of an internal three-fold cross-validation of the 128 training images.
In consensus with~\cite{Scharr2016}, we report the symmetric best Dice (SBD) and the absolute difference in count ($|$DiC$|$) and compare to other methods in Table~\ref{tbl:plant}.

\begin{figure}[t]
\begin{minipage}{0.5\textwidth}
\centering
\refstepcounter{table}
\label{tbl:plant_heart}
\subfloat[][Heart MRI input and segmentation.]
{\label{fig:heart}
\includegraphics[width=0.45\textwidth]{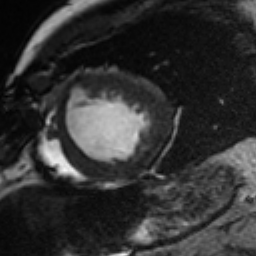}\hspace{1em}
\includegraphics[width=0.45\textwidth]{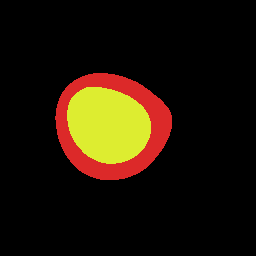}}\\
\subfloat[][Plant leaves input and instances.]{
\label{fig:plant}
\includegraphics[width=0.45\textwidth]{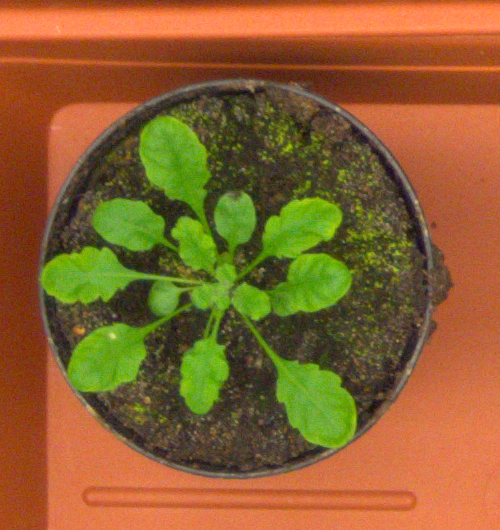}\hspace{1em}
\includegraphics[width=0.45\textwidth]{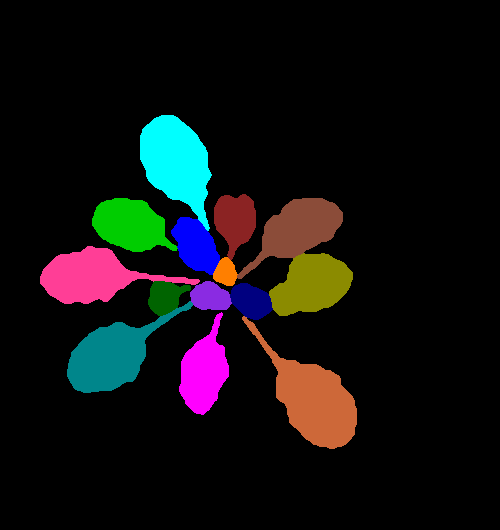}}
\addtocounter{figure}{-2}
\end{minipage}
\begin{minipage}{0.5\textwidth}
\centering
\captionsetup{position=top}
\refstepcounter{figure}
\subfloat[][Quantitative results of the heart MRI left ventricle segmentation.]{
\label{tbl:heart}
\begin{tabular}{c|c|c}
& IoU$_{\text{myo}}$ & IoU$_{\text{cav}}$\\
\hline
non-recurrent & $78.3 \pm 9.2$ & $89.1 \pm 7.7$\\
recurrent & $\textbf{79.4} \pm 8.5$ & $\textbf{89.4} \pm 7.2$\\
\end{tabular}}\\
\subfloat[][Quantitative results of the CVPPP leaf instance segmentation. Values taken from~\cite{Scharr2016}.]{
\label{tbl:plant}
\begin{tabular}{c|c|c}
& SBD & $|$DiC$|$\\
\hline
RIS+CRF & $66.6 \pm 8.7$ & $1.1 \pm 0.9$\\
MSU & $66.7 \pm 7.6$ & $2.3 \pm 1.6$\\
Nottingham & $68.3 \pm 6.3$ & $3.8 \pm 2.0$\\
Wageningen & $71.1 \pm 6.2$ & $2.2 \pm 1.6$\\
IPK & $74.4 \pm 4.3$ & $2.6 \pm 1.8$\\
IS+RA~\cite{Ren2017} & $\textbf{84.9} \pm 4.8$ & $\textbf{0.8} \pm 1.0$\\
\hline
Ours & $84.5 \pm 5.5$ & $1.5 \pm 1.2$\\
\end{tabular}}
\end{minipage}
\addtocounter{figure}{2}
\captionsetup{labelformat=andfigure}
\addtocounter{figure}{-1}
\caption{Results of the left ventricle segmentation and the CVPPP leaf instance segmentation.
Values show mean $\pm$ standard deviation.
Note that we report our results for both datasets based on a three-fold cross-validation setup.
Thus, they are not directly comparable to other published results.
SBD: symmetric best Dice; $|$DiC$|$: absolute difference in count; IoU: intersection over union; myo: myocardium; cav: blood cavity.}
\label{fig:plant_heart}
\end{figure}

\noindent\textbf{Cell Instance Tracking:}
As our main experiment, we show applicability of our full framework for instance segmentation and tracking by evaluating six different datasets of cell microscopy videos from the ISBI celltracking challenge~\cite{Ulman2017}.
Each celltracking dataset consists of two annotated training videos and two testing videos with image sizes ranging from $512\times512$ to $1200\times1024$ and with 48 to 138 frames.
We refer to~\cite{Maska2014} for additional imaging and video parameters.
As the instance IDs in groundtruth images are consistent throughout the whole video only for tracking, but not for segmentation, we merge both tracking and segmentation groundtruth for each frame to have consistent instance IDs.
Furthermore to learn the background embeddings, we only use the frames on which every cell is segmented.
With hyperparameters determined on the two annotated training videos from each dataset, we train the networks for predicting embeddings of size 16 on both videos for our challenge submission.

To compete in the tracking metric of the challenge, the framework is required to identify the parent ID of each cell.
As the framework is able to identify splitting cells and to assign new instance IDs (i.e., mitosis as seen on Fig.~\ref{fig:overview}), the parent ID of each newly created instance is determined as the instance with the highest IoU in previous frames.
We further postprocess the cells' family tree to be consistent with the evaluation criteria, e.g., an instance ID may not be used after splitting into children.
The results in comparison to the top performing methods are presented in Table~\ref{tbl:celltracking}.


\definecolor{ours}{RGB}{251, 118, 104}
\definecolor{fr_ro_ge}{RGB}{255,255,179}
\definecolor{kth_se_1_4}{RGB}{210,206,238}
\definecolor{cvut_cz}{RGB}{168,217,251}
\definecolor{bgu_il}{RGB}{255,220,138}
\definecolor{leid_nl}{RGB}{209,252,135}
\definecolor{cuni_cz}{RGB}{252,205,229
}
\definecolor{kit_ge}{RGB}{237,237,237}
\definecolor{hd_har_ge}{RGB}{233,173,234}
\definecolor{fr_be_ge}{RGB}{174,245,187}

\begin{table}[t]
\centering
\caption{Quantitative results of the celltracking datasets for overall performance~(OP), segmentation~(SEG), and tracking~(TRA), as described in~\cite{Ulman2017}.}
  \resizebox{\textwidth}{!}{%
\begin{tabular}{c p{0.4cm} | r | r | r | r | r | r | p{0.3cm} c}
 & & & & & & & & & \\
 & & \begin{tabular}[c]{@{}l@{}}DIC-\\ HeLa\end{tabular}& \begin{tabular}[c]{@{}l@{}}Fluo-\\ MSC\end{tabular} & \begin{tabular}[c]{@{}l@{}}Fluo-\\ GOWT1\end{tabular} & \begin{tabular}[c]{@{}l@{}}Fluo-\\ HeLa\end{tabular} & \begin{tabular}[c]{@{}l@{}}PhC-\\ U373\end{tabular} & \begin{tabular}[c]{@{}l@{}}Fluo-\\ SIM+\end{tabular} \\
\hhline{--|-|-|-|-|-|-}
\multirow{ 4}{*}{OP} & \nth{1} & \cellcolor{ours} \hphantom{xx} $0.864$ & \cellcolor{bgu_il} $0.759$ & \cellcolor{kth_se_1_4} $0.951$ & \cellcolor{kth_se_1_4} $0.942$ & \cellcolor{fr_ro_ge} $0.951$ & \cellcolor{bgu_il} $0.882$ &  & \cellcolor{ours} Ours \\
 & \nth{2} & \cellcolor{fr_ro_ge} $0.828$ & \cellcolor{kth_se_1_4} $0.676$ & \cellcolor{ours} $0.914$ & \cellcolor{fr_ro_ge} $0.940$ & \cellcolor{fr_be_ge} $0.896$ & \cellcolor{fr_ro_ge} $0.878$ &  & \cellcolor{bgu_il} BGU-IL (1--2)  \\
 & \nth{3} & \cellcolor{kth_se_1_4} $0.629$ & \cellcolor{cvut_cz} $0.658$ & \cellcolor{leid_nl} $0.902$ & \cellcolor{cvut_cz} $0.928$ & \cellcolor{cvut_cz} $0.895$ & \cellcolor{kth_se_1_4} $0.874$ &  & \cellcolor{cuni_cz} CUNI-CZ \\\hhline{~|~|-|-|-|-|-|-}
& & & \cellcolor{ours} \nth{5} $0.631$ & & \cellcolor{ours} \nth{11} $0.829$ & \cellcolor{ours} \nth{4} $0.888$ & \cellcolor{ours} \nth{9} $0.810$ &  & \cellcolor{cvut_cz} CVUT-CZ \\
\hhline{==|=|=|=|=|=|=}
\multirow{ 4}{*}{SEG} & \nth{1} & \cellcolor{ours} $0.814$ & \cellcolor{bgu_il} $0.645$ & \cellcolor{kth_se_1_4} $0.927$ & \cellcolor{fr_ro_ge} $0.903$ & \cellcolor{fr_ro_ge} $0.920$ & \cellcolor{bgu_il} $0.802$ &  & \cellcolor{fr_be_ge} FR-Be-GE \\
 & \nth{2} & \cellcolor{fr_ro_ge} $0.776$ & \cellcolor{kth_se_1_4} $0.590$ & \cellcolor{leid_nl} $0.893$ & \cellcolor{kth_se_1_4} $0.893$ & \cellcolor{cvut_cz} $0.832$ & \cellcolor{kth_se_1_4} $0.791$ &  & \cellcolor{fr_ro_ge} FR-Ro-GE \\
 & \nth{3} & \cellcolor{cvut_cz} $0.464$ & \cellcolor{fr_ro_ge} $0.582$ & \cellcolor{cuni_cz} $0.887$ & \cellcolor{cvut_cz} $0.869$ & \cellcolor{fr_be_ge} $0.826$ & \cellcolor{fr_ro_ge} $0.781$ &  & \cellcolor{hd_har_ge} HD-Har-GE \\\hhline{~|~|-|-|-|-|-|-}
& & & \cellcolor{ours} \nth{5} $0.496$ & \cellcolor{ours} \nth{4} $0.880$ & \cellcolor{ours} \nth{10} $0.749$ & \cellcolor{ours} \nth{5} $0.793$ & \cellcolor{ours} \nth{8} $0.718$ &  & \cellcolor{kit_ge} KIT-GE \\
\hhline{==|=|=|=|=|=|=}
\multirow{ 4}{*}{TRA} & \nth{1} & \cellcolor{ours} $0.915$ & \cellcolor{bgu_il} $0.873$ & \cellcolor{kth_se_1_4} $0.976$ & \cellcolor{kth_se_1_4} $0.991$ & \cellcolor{ours} $0.983$ & \cellcolor{fr_ro_ge} $0.975$ &  & \cellcolor{kth_se_1_4} KTH-SE (1--4) \\
 & \nth{2} & \cellcolor{fr_ro_ge} $0.881$ & \cellcolor{ours} $0.765$ & \cellcolor{ours} $0.947$ & \cellcolor{cvut_cz} $0.987$ & \cellcolor{fr_ro_ge} $0.981$ & \cellcolor{bgu_il} $0.961$ &  & \cellcolor{leid_nl} LEID-NL \\
 & \nth{3} & \cellcolor{kth_se_1_4} $0.797$ & \cellcolor{kth_se_1_4} $0.763$ & \cellcolor{kit_ge} $0.925$ & \cellcolor{hd_har_ge} $0.986$ & \cellcolor{kth_se_1_4} $0.977$ & \cellcolor{kth_se_1_4} $0.957$ &  &  \\\hhline{~|~|-|-|-|-|-|-}
& & & &  & \cellcolor{ours} \nth{12} $0.909$ & & \cellcolor{ours} \nth{10} $0.902$ &  & \\
\hhline{--|-|-|-|-|-|-}
\end{tabular}
}
\label{tbl:celltracking}
\end{table}

\section{Discussion and Conclusion}
\label{sec:conclusion}

Up to our knowledge, we are the first to present a method that incorporates temporal information into a network to allow tracking of embeddings for instance segmentation.
We perform three experiments to show different aspects of our novel method, i.e., temporal segmentation, instance segmentation, and combined instance segmentation and tracking. 
Thus, we demonstrate the wide applicability of our approach.

We use the left ventricle segmentation experiment to show that our novel recurrent stacked hourglass network can be used for incorporating temporal information.
It can be seen from the results of the experiment that incorporating ConvGRU between contracting and expanding path deeply inside the architecture improves over the baseline stacked hourglass network.
Nevertheless, since we simplified the evaluation protocol of the challenge, the results of the experiment should not be directly compared to other reported results.
Moreover, benefits of such deep incorporation compared to having recurrent layers on other positions in the network~\cite{Ren2017} remain to be shown.

This paper also contributes with a novel embedding loss based on cosine similarities.
Most of the methods that use embeddings for differentiating between instance segmentations are based on maximizing distances of embeddings in the Euclidean space, e.g.,~\cite{Newell2017}.
When using such embedding losses, we observed problems when combining them with recurrent networks, presumably due to unrestricted embedding values.
To overcome these problems, we use cosine similarities that normalize embeddings.
The only other work that suggests cosine similarities for instance segmentation with embeddings is the unpublished work of~\cite{Kong2017}.
However, compared to their embedding loss that takes all instances into account, our novel loss focuses only on neighboring ones, which can be beneficial for optimization in the case of a large number of instances.
We evaluate our novel loss on the CVPPP challenge dedicated to instance segmentation from still images.
While waiting for the results of the competition, our method evaluated with three-fold cross-validation shows to be in line with the currently leading method, and has a significant margin to the second best.
Moreover, compared to the leading method~\cite{Ren2017}, the architecture of our method is considerably simpler. 




In our main experiment for segmentation and tracking of instances, we evaluate our method on the ISBI celltracking challenge, showing large variability in visual appearance, size and number of cells.
Our method achieves two first and two second places among the six submitted datasets in the tracking metric.
For the dataset DIC-HeLa, having a dense layout of cells as seen in Fig.~\ref{fig:overview}, we outperform all other methods in both tracking and segmentation metrics. 
On the dataset Fluo-GOWT1 we rank overall second.
On the datasets Fluo-HeLa and Flou-SIM+, which consist of images with small cells, our method does not perform well due to  the need to downsample images for the network to process them.
When the downsampling results in drastic reduction of cell sizes, our method fails to create instance segmentations, 
thus explaining the not satisfying performance also in tracking.
To increase the resolution and consequently improve segmentation and tracking, we could split the input image into multiple smaller parts, similarly as done in~\cite{Ronneberger2015}.

In conclusion, our work has shown that embeddings for instance segmentation can be successfully combined with recurrent networks incorporating temporal information to perform instance tracking.
In future work, we will investigate the possibility of incorporating the required clustering step inside of a single end-to-end trained network, which could simplify the framework and further improve the segmentation and tracking results.

\bibliographystyle{splncs03}
\bibliography{main}

\begin{appendices}
\renewcommand{\thesection}{\arabic{section}}%

\nociteappendix{*}

\chapter*{Appendix}

\section{Network and Training Parameters}
We set the network parameters as follows:
The weights of each convolution layer of the stacked hourglass network are initialized with the method as described in~\cite{He2015}, the biases with~0.
The networks do not employ any normalization layers or dropout, but use an L2 weight regularization factor of 0.00001.
Due to the demanding training of recurrent neural networks, in terms of both memory and computational requirements, we set the mini-batch size to 1.
We train the recurrent networks for sequences of 10 consecutive frames.
For the non-recurrent neural networks, we use a mini-batch size of 10.
We train all networks with ADAM~\cite{Kingma2015} for total 40000 iterations and a learning rate of 0.0001, while the learning rate is reduced to 0.00001 after 20000 iterations.
Training of a recurrent networks took $\approx12$ hours, training of the non-recurrent networks took $\approx8$ hours on a single NVIDIA Titan Xp with 12 GB.

\section{Data Preprocessing and Augmentation Parameters}

We perform input data augmentation, by changing intensity values and spatial deformations.
First, we change the image intensity values such that the minimum and maximum values are $-1$ and $1$.
As MR and microscopy images may contain outliers in terms of minimum and maximum values, we calculate the minimum value as the median of $i_{\text{min}}$\% of all intensity values of an image, and the maximum as the median of $i_{\text{max}}$\%.
Then, for augmentation, we shift each intensity value randomly by $i_{\text{shift}}$ and scale each intensity by $i_{\text{scale}}$.
For the random spatial deformations in both $x$ and $y$ axes, we translate by $t$ pixels, flip axis with probability $f_p$, rotate by $r$ degrees and scale by $s$.
Furthermore, we employ elastic deformations, by randomly moving points by $b$ pixels on a grid of size $g$ and interpolating with third order splines.
All random augmentations sample from a uniform distribution within the specified intervals.

\noindent\textbf{Left Ventricle Segmentation:}
The augmentation parameters are as follows:
Intensity transformations: $i_{\text{min}}=10\%$, $i_{\text{max}}=10\%$, $i_{\text{shift}} \in [-0.25, 0.25]$, $i_{\text{scale}}\in[0.75, 1.25]$.
Spatial transformations: $t\in[-20, 20]$, $f_p=0$, $r\in[-15, 15]$, $s\in[0.75, 1.25]$, $b=8$, $g\in[-10, 10]$.
We set default pixel values outside the defined image region to 0.

\noindent\textbf{Leaf Instance Segmentation:}
The augmentation parameters are as follows:
Intensity transformations: $i_{\text{min}}=1\%$, $i_{\text{max}}=1\%$, $i_{\text{shift}} \in [-0.25, 0.25]$, $i_{\text{scale}}\in[0.75, 1.25]$.
Spatial transformations: $t\in[-12, 12]$, $f_p=0.5$, $r\in[-180, 180]$, $s\in[0.75, 1.25]$, $b=8$, $g\in[-10, 10]$.
For each instance $i\in\mathbb{I}$, we define all pixels inside the segmentation mask as $\mathbb{S}^{(i)}$, and all pixels of all other instances as $\mathbb{N}^{(i)}$.
We perform mirror padding for pixels outside the defined image region, but we do not calculate the loss for these pixels.

\noindent\textbf{Cell Instance Tracking:}
Unless otherwise stated, the augmentation parameters for all datasets are as follows:
Intensity transformations: $i_{\text{min}}=20\%$, $i_{\text{max}}=10\%$ (for Fluo-MSC and Fluo-SIM+ we set $i_{\text{max}}=1\%$), $i_{\text{shift}} \in [-0.25, 0.25]$, $i_{\text{scale}}\in[0.75, 1.25]$.
Due to noise in the intensity values, we smooth the images with a Gaussian function with $\sigma=2$ pixel.
Spatial transformations: $t\in[-25, 25]$, $f_p=0.5$, $r\in[-180, 180]$, $s\in[0.75, 1.25]$, $b=8$, $g\in[-10, 10]$.
%
For each instance $i$, we define all pixels inside the segmentation mask as $\mathbb{S}^{(i)}$, while we set $\mathbb{N}^{(i)}$ to only neighboring instances within a specified radius $r_{\mathbb{N}}$ in pixels.
For dataset Fluo-MSC we set $r_{\mathbb{N}}=150$, for the dataset Fluo-HeLa we set $r_{\mathbb{N}}=25$. For all other datasets we set $r_{\mathbb{N}}=50$.
For each mini-batch, we use at most 32 different instances for training, to reduce memory consumption.
We perform mirror padding for pixels outside the defined image region, but we do not calculate the loss for these pixels.

\section{Clustering Parameters}

We append the image coordinates scaled with factor $c$ to value of the embeddings as data points for the clustering algorithm.
We modify the parameters $c$ and $m_{\text{pts}}$ for each dataset, while we set $m_{\text{clSize}}=m_{\text{pts}}$ and $t_{\text{size}}=\frac{m_{\text{pts}}}{2}$.
DIC-HeLa: $m_{\text{pts}}=1000$, $c=0.02$;
Fluo-MSC: $m_{\text{pts}}=500$, $c=0.1$;
Fluo-GOWT1: $m_{\text{pts}}=50$, $c=0.001$;
Fluo-SIM+: $m_{\text{pts}}=100$, $c=0.001$;
Fluo-HeLa: $m_{\text{pts}}=25$, $c=0.01$;
PhC-U373: $m_{\text{pts}}=500$, $c=0.005$;
For the CVPPP dataset we set $m_{\text{pts}}=50$, $c=0.001$.

\begin{figure}[htbp]
\begin{center}
\subfloat[][DIC-HeLa]
{\includegraphics[width=0.32\textwidth]{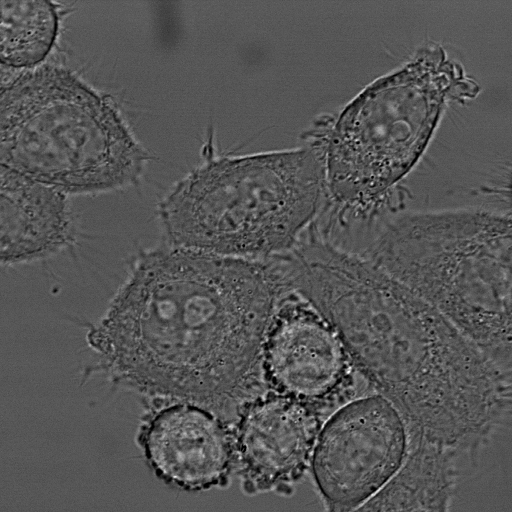}\hspace{1ex}
\includegraphics[width=0.32\textwidth]{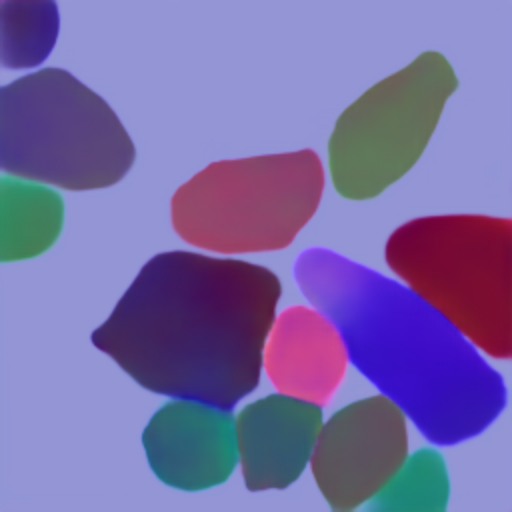}\hspace{1ex}
\includegraphics[width=0.32\textwidth]{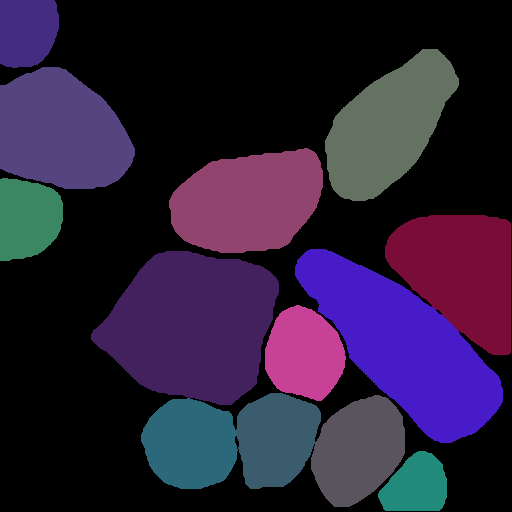}}\\
\subfloat[][Fluo-MSC]
{\includegraphics[width=0.32\textwidth]{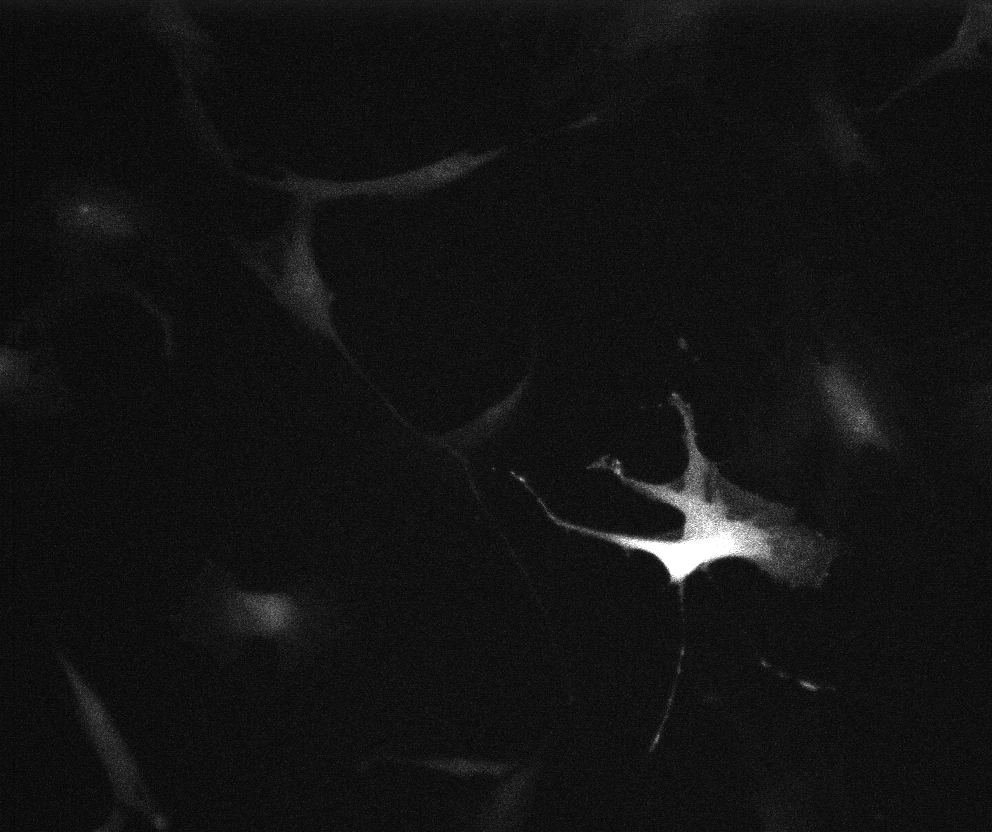}\hspace{1ex}
\includegraphics[width=0.32\textwidth]{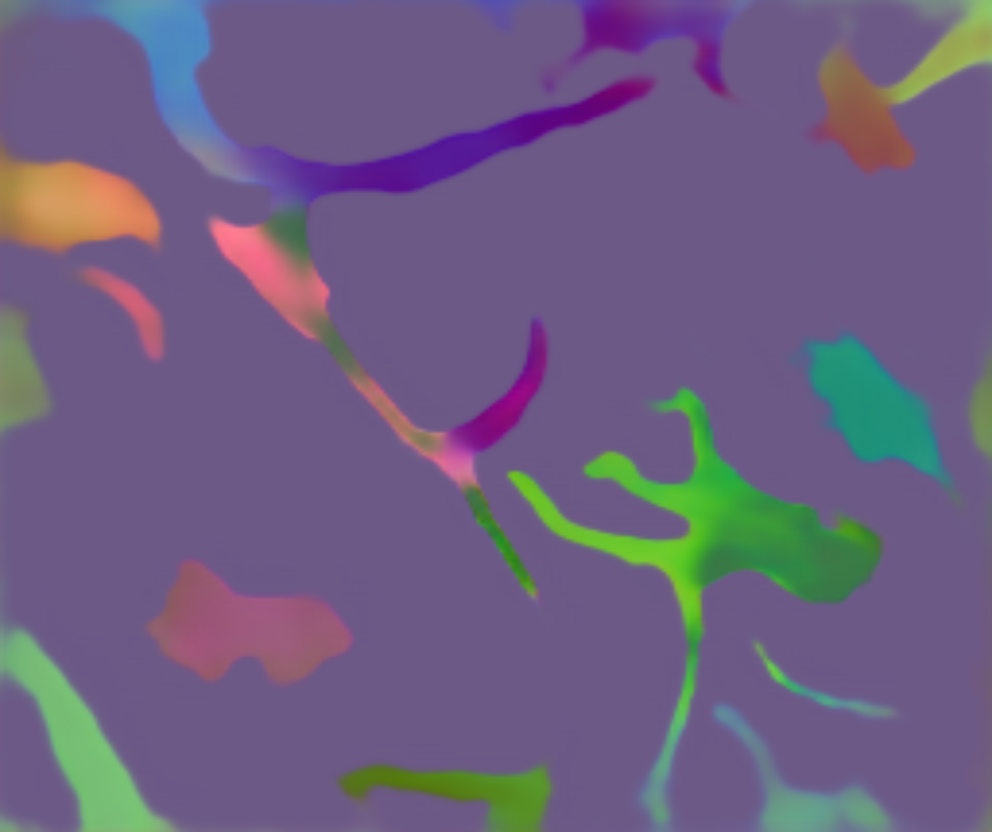}\hspace{1ex}
\includegraphics[width=0.32\textwidth]{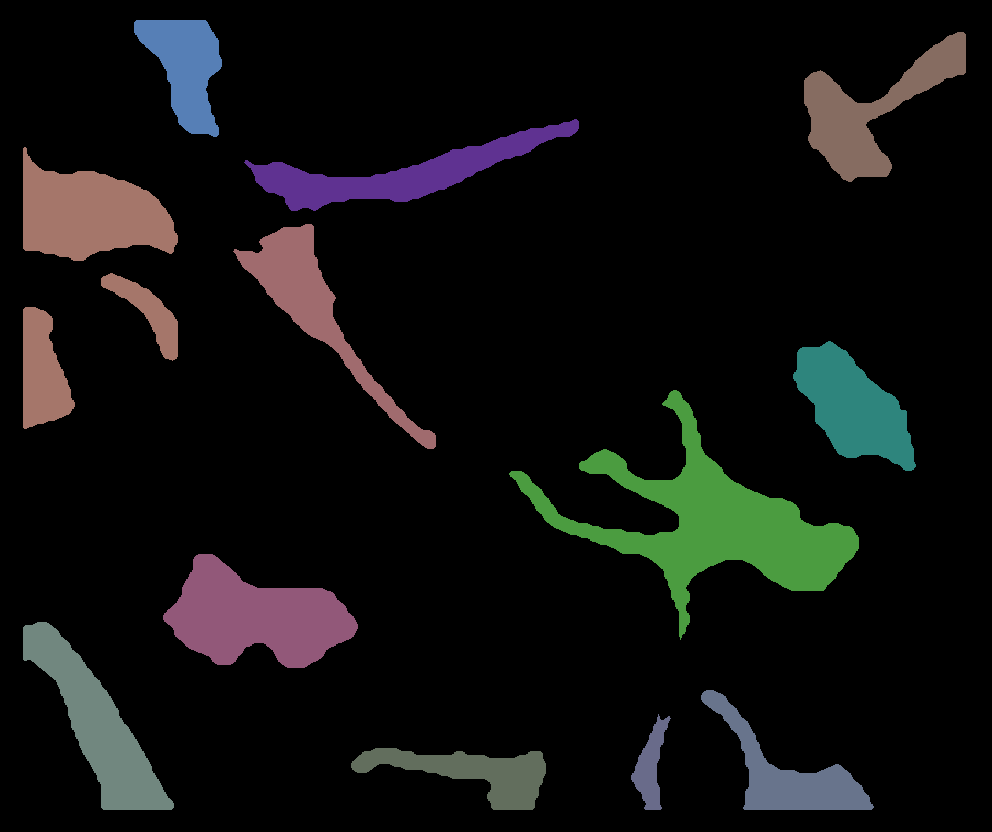}}\\
\subfloat[][Fluo-GOWT1]
{\includegraphics[width=0.32\textwidth]{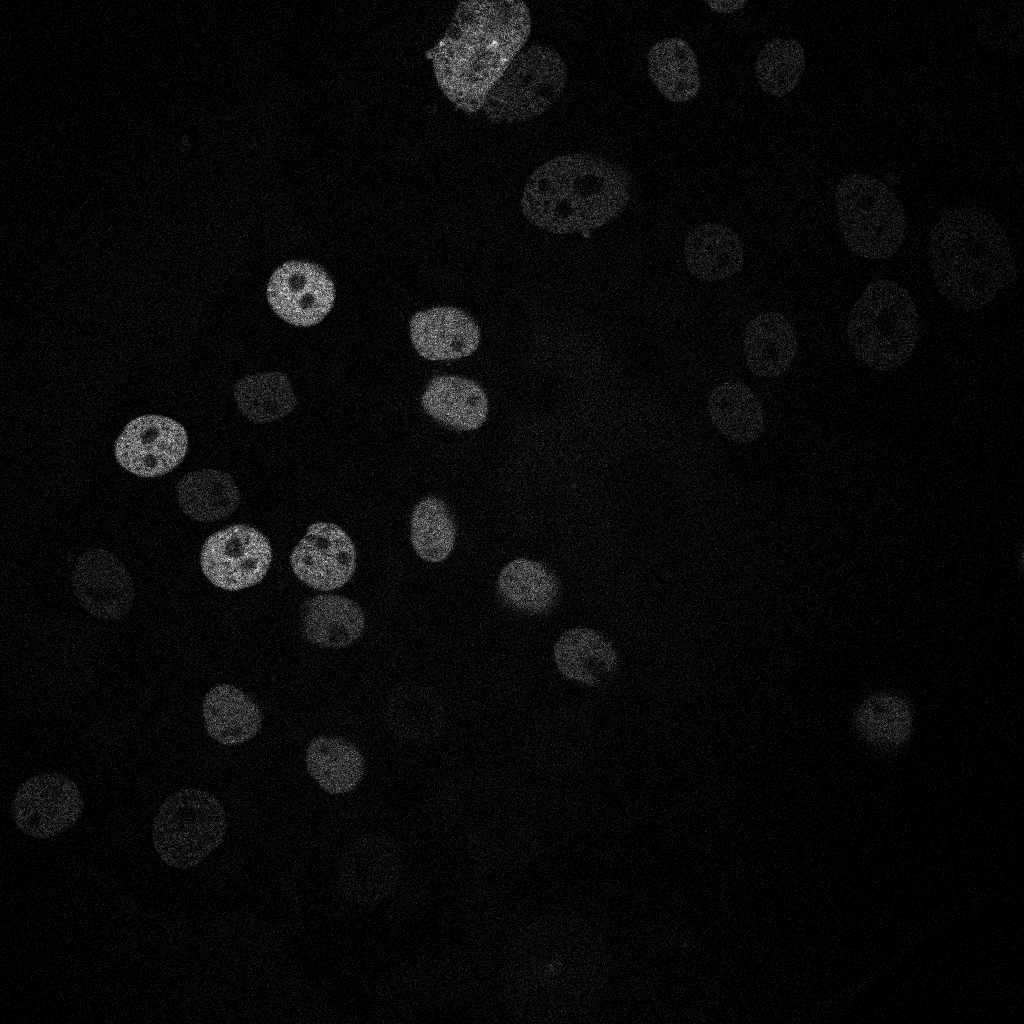}\hspace{1ex}
\includegraphics[width=0.32\textwidth]{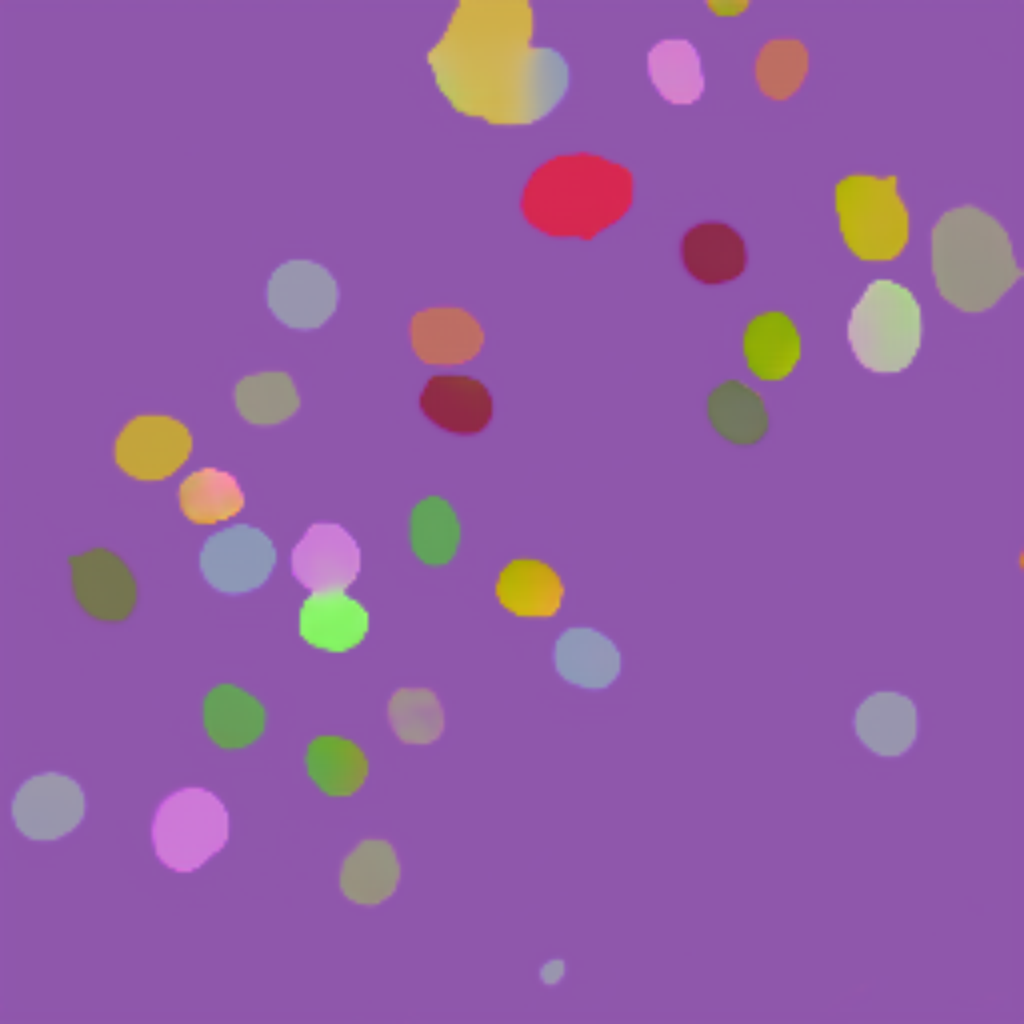}\hspace{1ex}
\includegraphics[width=0.32\textwidth]{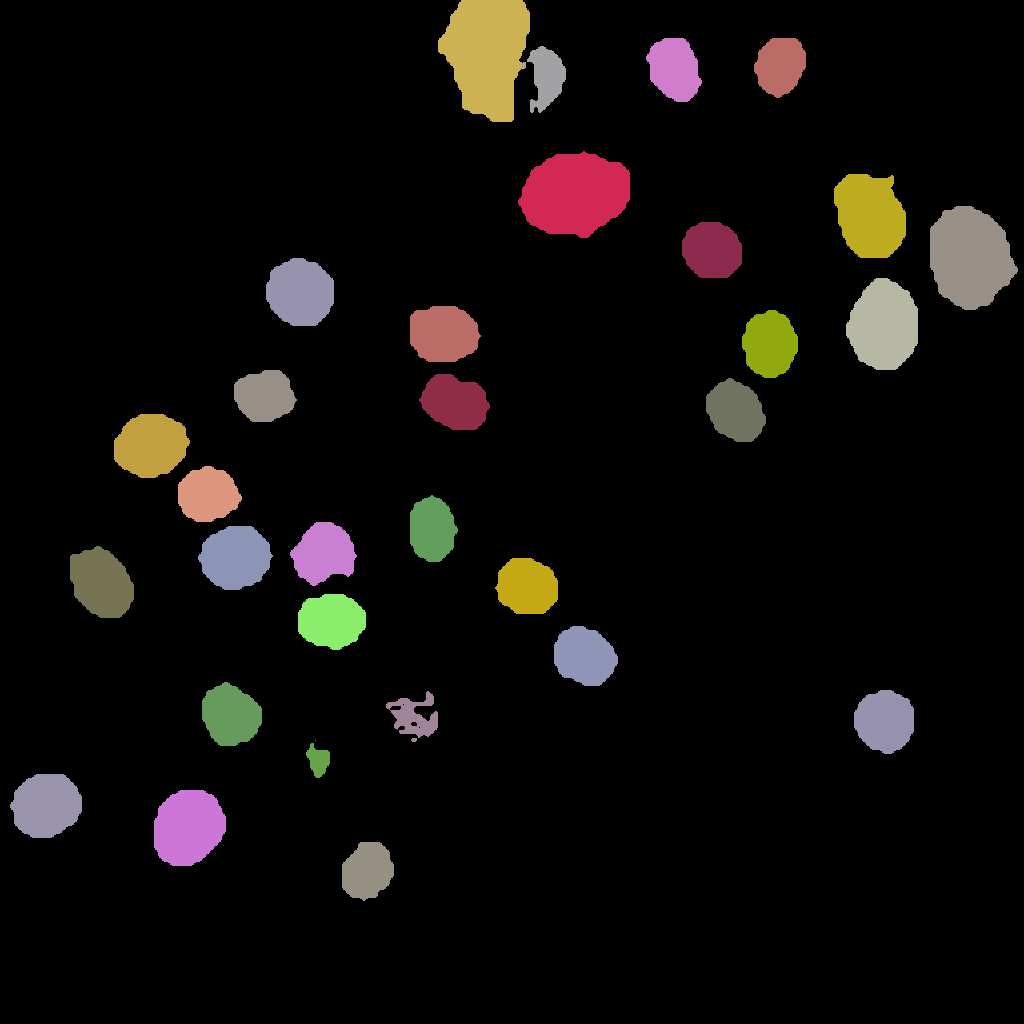}}
\caption{Example results of the evaluated celltracking datasets. Left: normalized input; middle: three randomly chosen dimensions of the embedding as RGB channels; right: final instance segmentation.}
\label{default}
\end{center}
\end{figure}

\begin{figure}[htbp]
\begin{center}
\subfloat[][Fluo-SIM+]
{\includegraphics[width=0.32\textwidth]{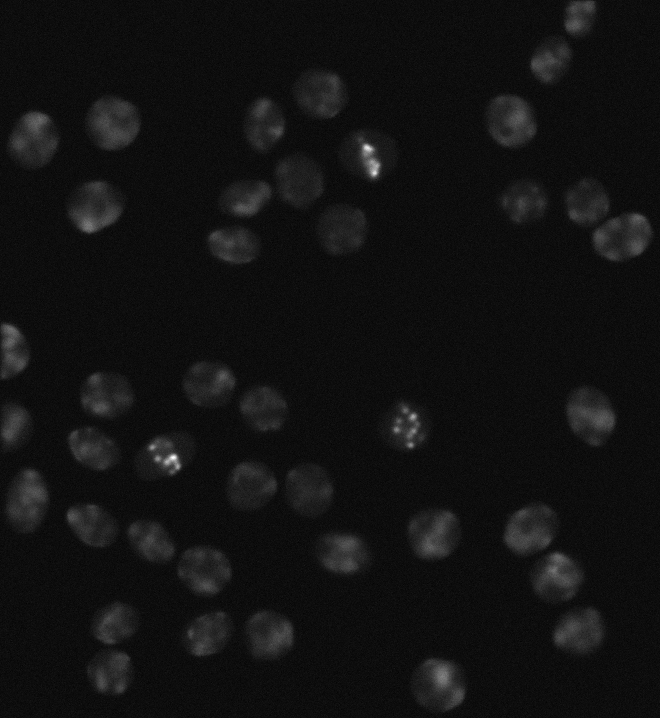}\hspace{1ex}
\includegraphics[width=0.32\textwidth]{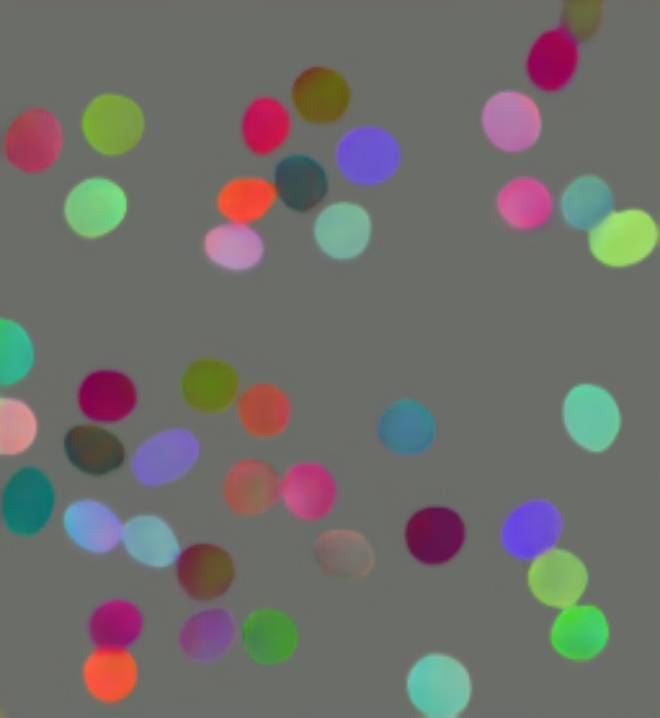}\hspace{1ex}
\includegraphics[width=0.32\textwidth]{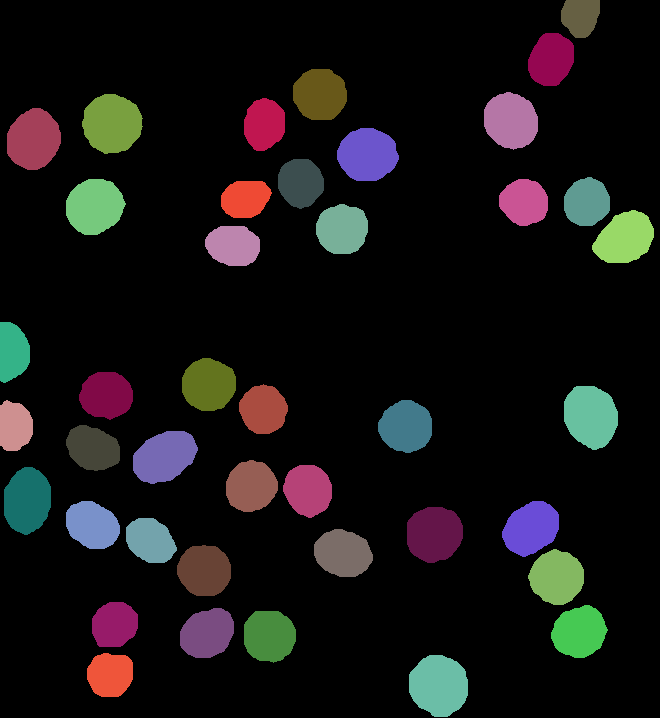}}\\
\subfloat[][Fluo-HeLa]
{\includegraphics[width=0.32\textwidth]{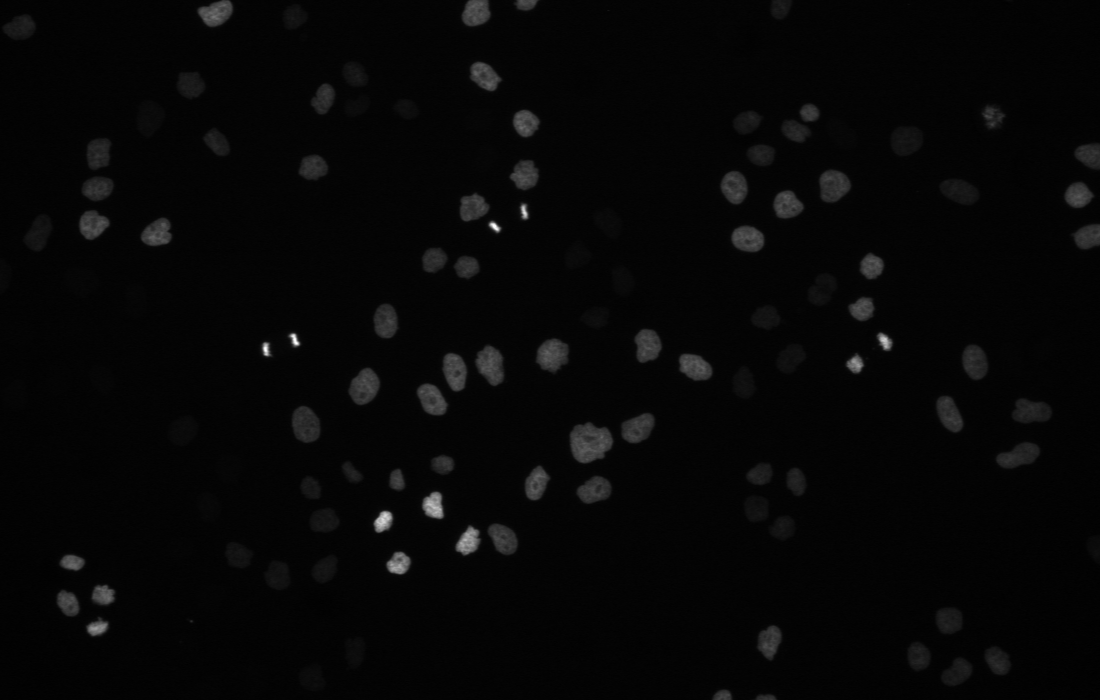}\hspace{1ex}
\includegraphics[width=0.32\textwidth]{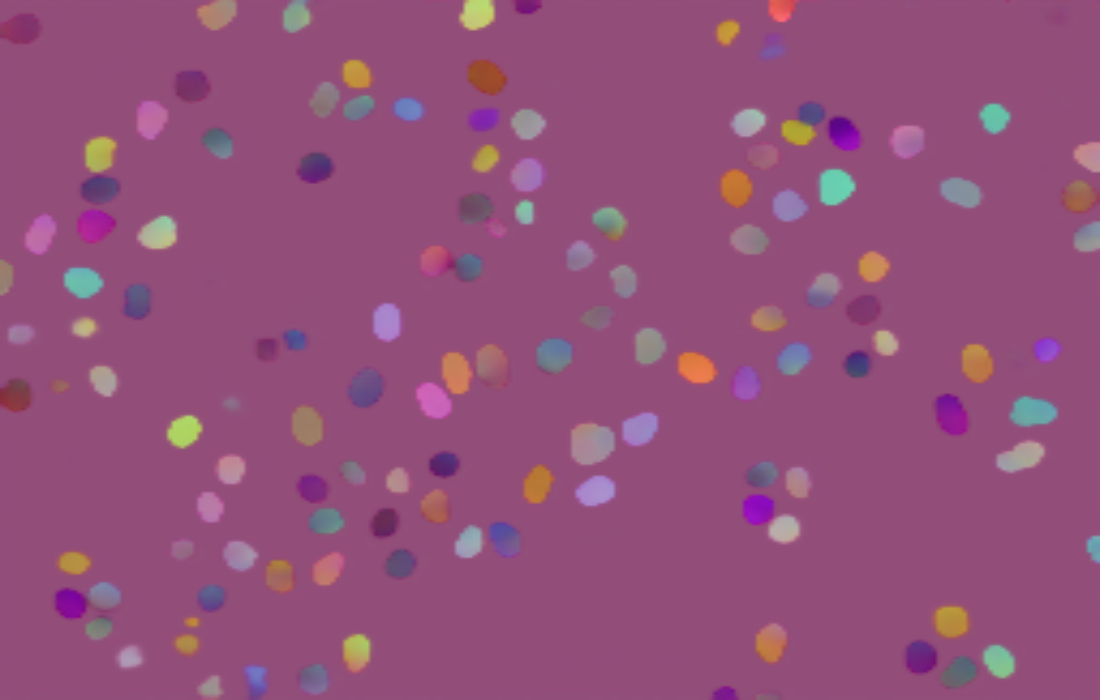}\hspace{1ex}
\includegraphics[width=0.32\textwidth]{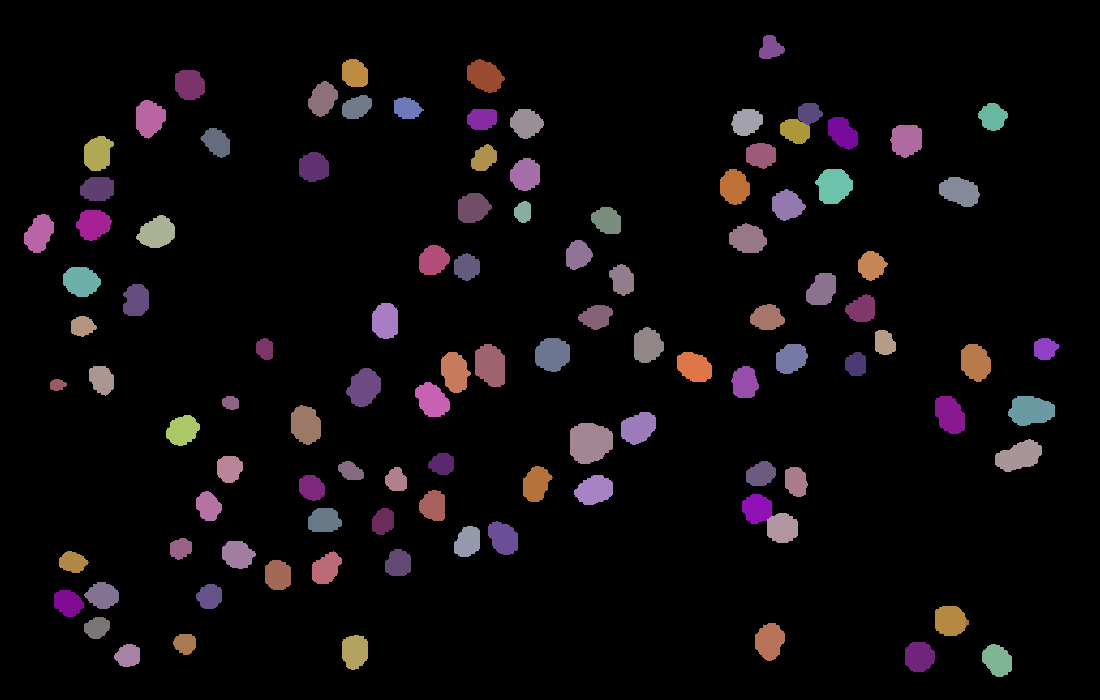}}\\
\subfloat[][PhC-U373]
{\includegraphics[width=0.32\textwidth]{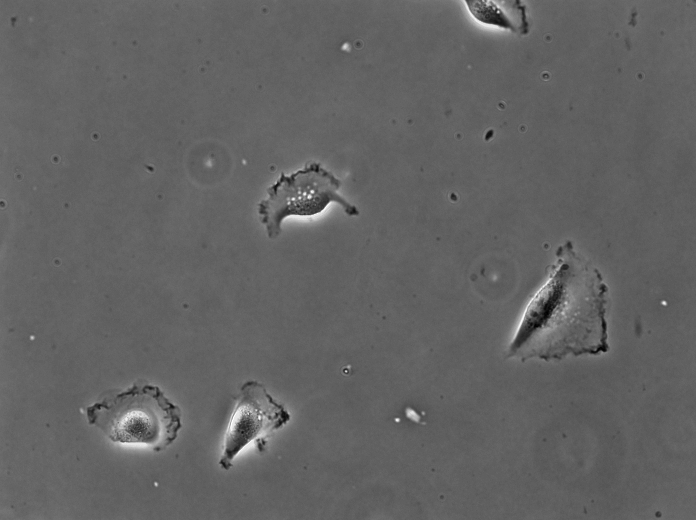}\hspace{1ex}
\includegraphics[width=0.32\textwidth]{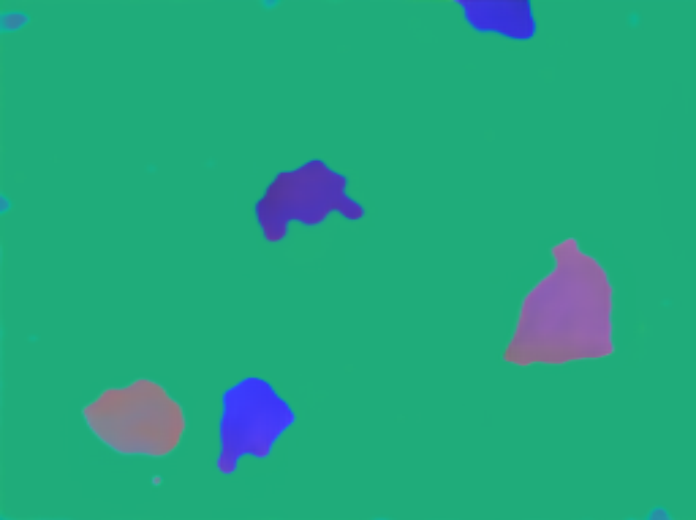}\hspace{1ex}
\includegraphics[width=0.32\textwidth]{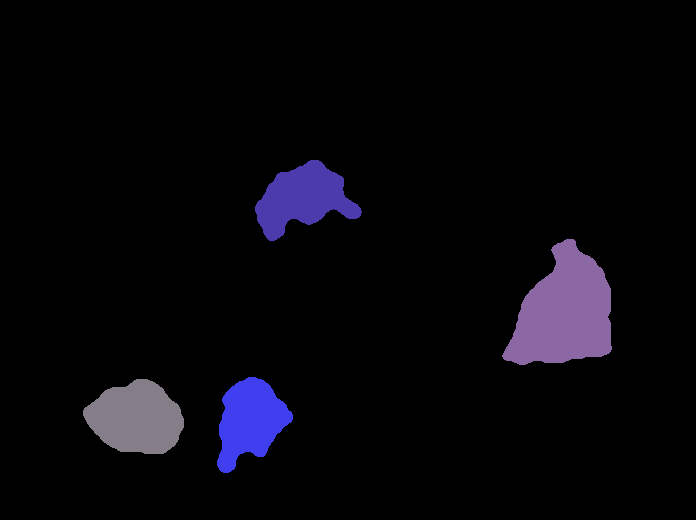}}
\caption{Example results of the evaluated celltracking datasets, continued.}
\label{default}
\end{center}
\end{figure}

\end{appendices}

\bibliographystyleappendix{splncs03}
\bibliographyappendix{appendix}

\end{document}